\documentclass[sigconf]{acmart}

\AtBeginDocument{%
  }

\copyrightyear{2025}
\acmYear{2025}
\setcopyright{cc}
\setcctype{by}
\acmConference[WWW '25]{Proceedings of the ACM Web Conference 2025}{April 28-May 2, 2025}{Sydney, NSW, Australia}
\acmBooktitle{Proceedings of the ACM Web Conference 2025 (WWW '25), April 28-May 2, 2025, Sydney, NSW, Australia}
\acmDOI{10.1145/3696410.3714516}
\acmISBN{979-8-4007-1274-6/25/04}

\usepackage[utf8]{inputenc} 
\usepackage[T1]{fontenc}    
\usepackage{hyperref}       
\usepackage{url}            
\usepackage{booktabs}       
\usepackage{amsfonts}       
\usepackage{nicefrac}       
\usepackage{microtype}      
\usepackage{xcolor}         
\usepackage{tabularx}
\usepackage{pifont}
\usepackage{array}
\usepackage{rotating}
\usepackage{wrapfig}
\usepackage{float}
\usepackage{microtype}
\usepackage{graphicx}
\usepackage{booktabs} 
\usepackage{caption}
\usepackage{multirow}
\usepackage{ulem}
\usepackage{enumitem}
\usepackage{amsmath}
\usepackage[noend]{algorithmic}
\usepackage{algorithm}
\usepackage{algorithmic}
\usepackage{subcaption}
\usepackage{bm}
\usepackage{marvosym}

\newcolumntype{Y}{>{\centering\arraybackslash}X}
\newcommand{\cxmark}{\ding{52}\rotatebox[origin=c]{-9.2}{\kern-0.7em\ding{55}}}%

\begin{document}

\title{Noise Matters: Diffusion Model-based Urban Mobility Generation with Collaborative Noise Priors}

\author{Yuheng~Zhang}
\affiliation{
\institution{Department~of~Electronic~Engineering, BNRist, Tsinghua University}
\city{Beijing}
\country{China}
}
\email{yuheng-z22@mails.tsinghua.edu.cn}
\authornotemark[1]

\author{Yuan~Yuan}
\affiliation{
\institution{Department~of~Electronic~Engineering, BNRist, Tsinghua University}
\city{Beijing}
\country{China}
}
\email{y-yuan20@mails.tsinghua.edu.cn}
\authornote{Equal contribution.}

\author{Jingtao~Ding}
\affiliation{
\institution{Department~of~Electronic~Engineering, BNRist, Tsinghua University}
\city{Beijing}
\country{China}
}
\email{dingjt15@tsinghua.org.cn}

\author{Jian~Yuan}
\affiliation{
\institution{Department~of~Electronic~Engineering, BNRist, Tsinghua University}
\city{Beijing}
\country{China}
}
\email{jyuan@tsinghua.edu.cn}

\author{Yong~Li}
\affiliation{
\institution{Department~of~Electronic~Engineering, BNRist, Tsinghua University}
\city{Beijing}
\country{China}
}
\email{liyong07@tsinghua.edu.cn}
\authornote{Corresponding author.}

\renewcommand{\shortauthors}{Yuheng Zhang, Yuan Yuan, Jingtao Ding, Jian Yuan, and Yong Li.}

\begin{abstract}
With global urbanization, the focus on sustainable cities has largely grown, driving research into equity, resilience, and urban planning, which often relies on mobility data. The rise of web-based apps and mobile devices has provided valuable user data for mobility-related research. However, real-world mobility data is costly and raises privacy concerns. To protect privacy while retaining key features of real-world movement, the demand for synthetic data has steadily increased. Recent advances in diffusion models have shown great potential for mobility trajectory generation due to their ability to model randomness and uncertainty. However, existing approaches often directly apply identically distributed (i.i.d.) noise sampling from image generation techniques, which fail to account for the spatiotemporal correlations and social interactions that shape urban mobility patterns. In this paper, we propose CoDiffMob, a diffusion model for urban mobility generation with collaborative noise priors, we emphasize the critical role of noise in diffusion models for generating mobility data. By leveraging both individual movement characteristics and population-wide dynamics, we construct novel collaborative noise priors that provide richer and more informative guidance throughout the generation process. Extensive experiments demonstrate the superiority of our method, with generated data accurately capturing both individual preferences and collective patterns, achieving an improvement of over 32\%. Furthermore, it can effectively replace web-derived mobility data to better support downstream applications, while safeguarding user privacy and fostering a more secure and ethical web. This highlights its tremendous potential for applications in sustainable city-related research. The code and data are available at \color{blue}\url{https://github.com/tsinghua-fib-lab/CoDiffMob}.
\end{abstract}



\maketitle

\section{Introduction}
The UN Sustainable Development Goals~\cite{un_sdg_report_2024} articulate a vision for developing inclusive, safe, resilient, and sustainable cities,   highlighting the importance of affordability, efficiency in public transportation, and sustainable land use management.
Research in these areas typically depends on city-level mobility data as an essential foundation~\cite{song2016deeptransport, zheng2014urban, zheng2009mining, zhang2025metacity}.
The widespread adoption of web-based applications and mobile devices has greatly expanded the capacity to collect user mobility data~\cite{hess2015data,yuan2023learning} while increasing the demand for utilizing and analyzing these data to further optimize web-related applications, such as location and service recommendation. However, direct usage of these real-world mobility data often raises serious privacy concerns~\cite{rao2020lstm_privacy,zhang2023dp_privacy}, and their collection is typically associated with high costs.
For the good of urban residents and sustainable development of web applications, there is a growing demand for generating synthetic urban mobility data~\cite{liu2024act2loc, yuan2022activity, shao2024beyond, yuan2024generating} that not only preserves the key characteristics of real-world movement for the utility of user-centric web services and urban decision-making~\cite{yuan2024foundation,zheng2024survey, feng2019identification} but also ensures that user privacy is well protected.

\begin{figure*}[ht]
\centering
\begin{subfigure}[ht]{1.0\linewidth}
    \centering
    \includegraphics[width=0.9\linewidth]{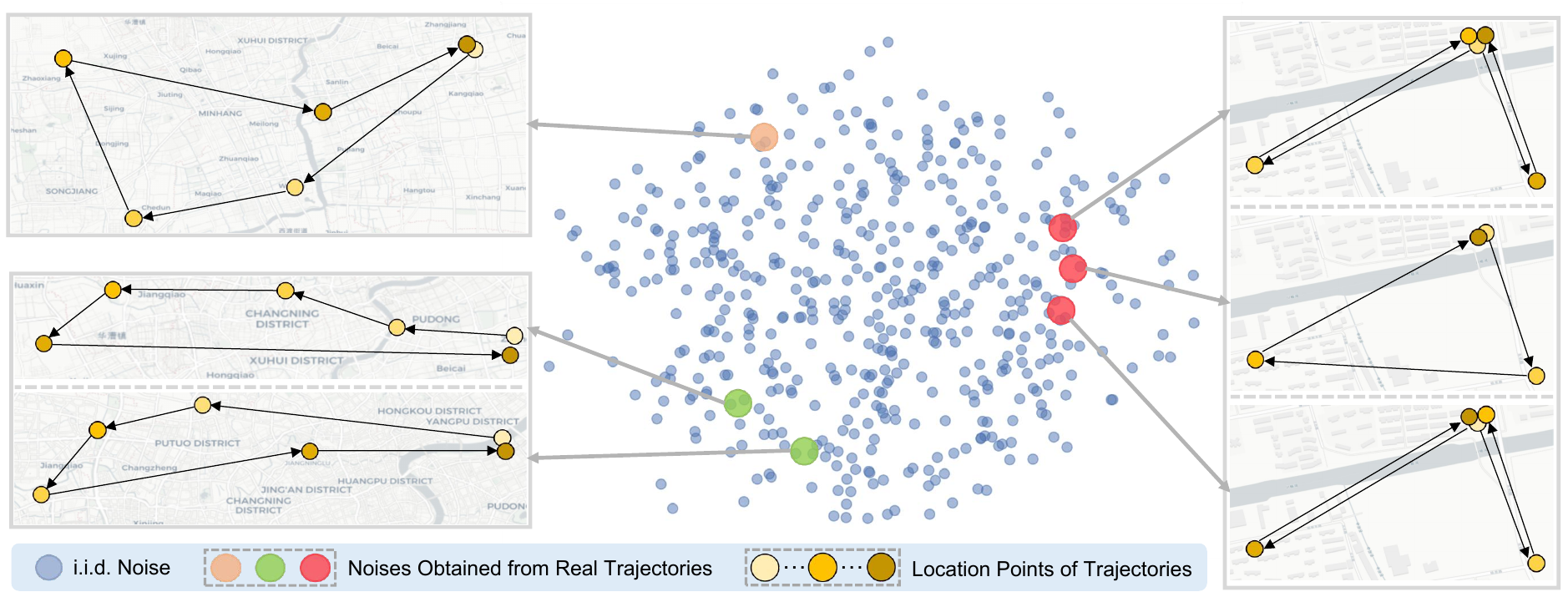}
    \caption{Noise obtained from real trajectories with different mobility patterns.}
    \label{fig:noise2traj}
\end{subfigure}
\begin{subfigure}[ht]{0.6\linewidth}
    \hspace{-0.6em}
    \includegraphics[width=0.5\linewidth]{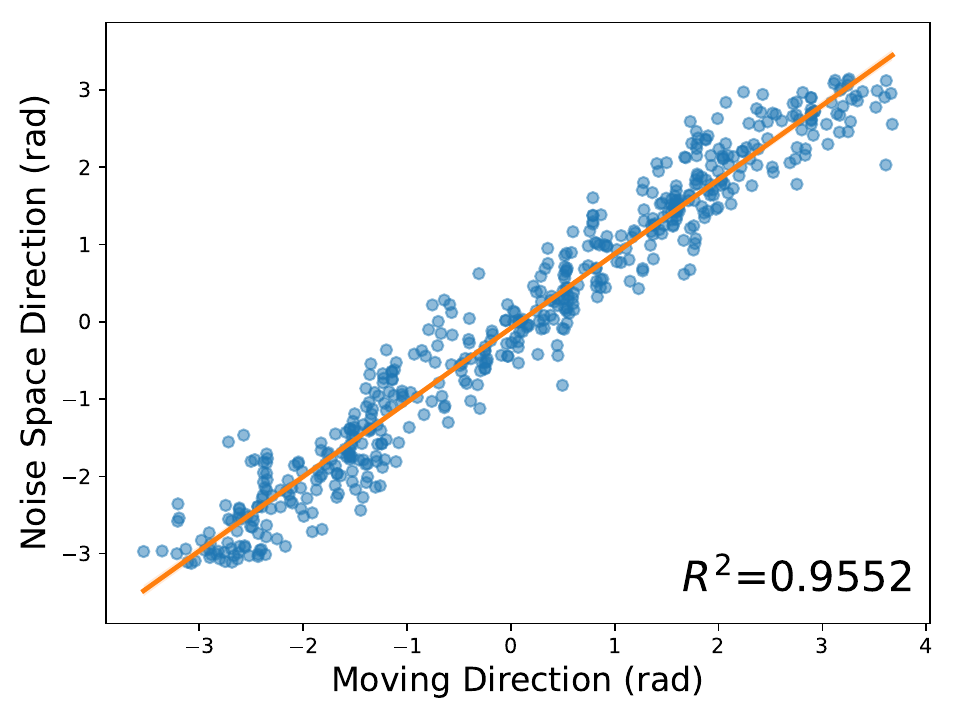}
    \includegraphics[width=0.5\linewidth]{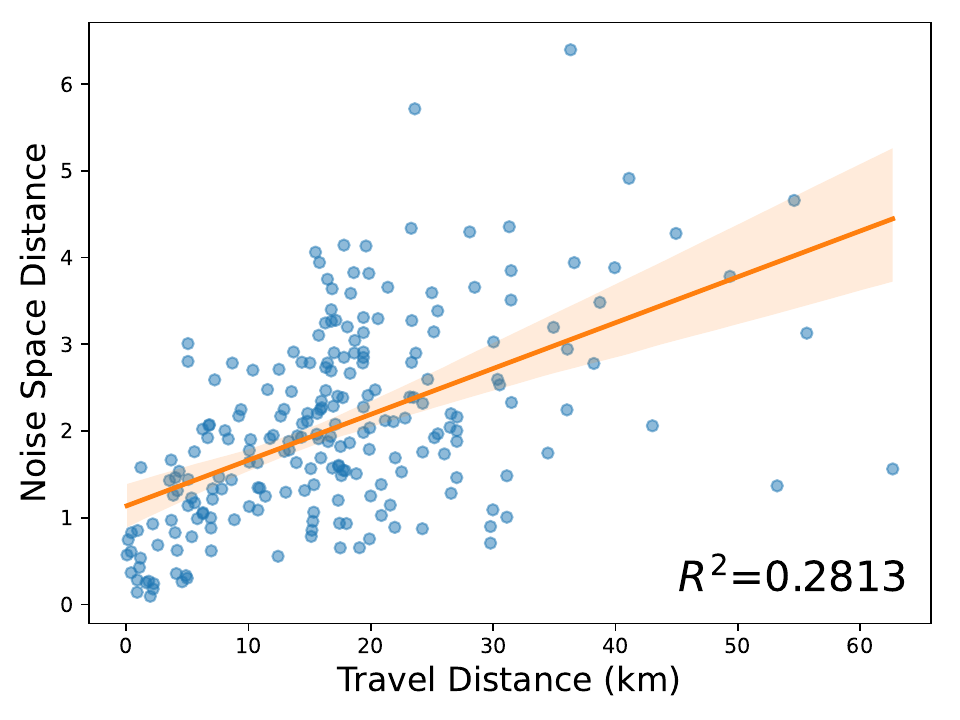}
    \caption{Correlation in the spatial domain.}
    \label{fig:corr_spatial}
\end{subfigure}
\begin{subfigure}[ht]{0.3\linewidth}
    \hspace{0.4em}
    \includegraphics[width=1\linewidth]{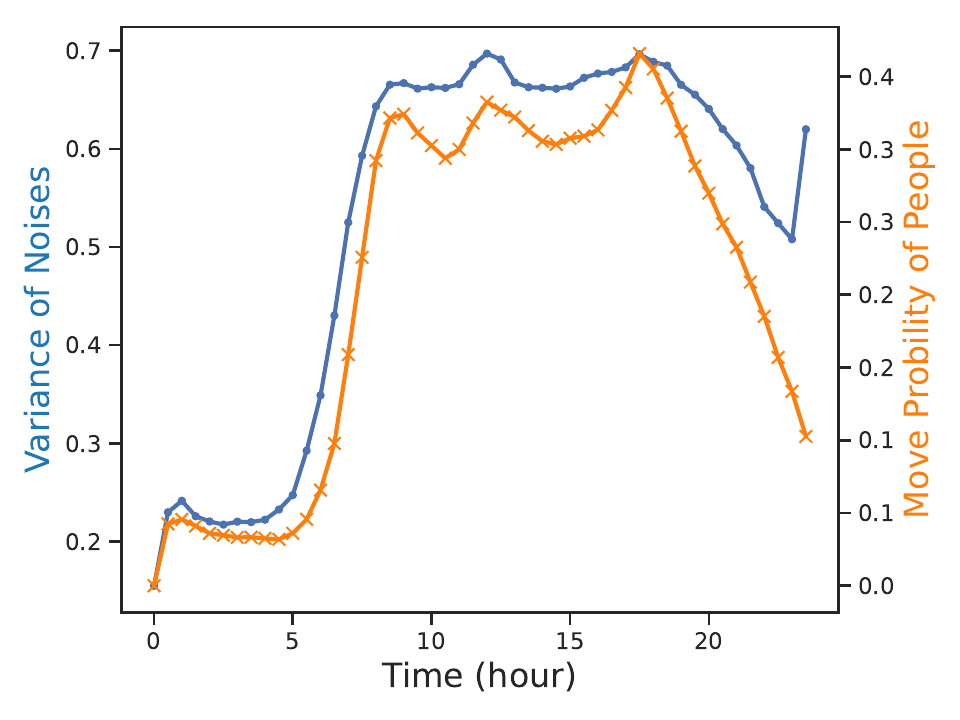}
    \caption{Correlation in the temporal domain.}
    \label{fig:corr_time}
\end{subfigure}
\vspace{-0.5em}
\caption{Correlations between noise and trajectories. (a) shows the t-SNE plot of the noise corresponding to the trajectories. (b) and (c) illustrate the correlations between the noise and trajectories from the spatial and temporal domain.}
\label{fig:noise_map}
\end{figure*}

Diffusion models~\cite{long2024universal,ho2020ddpm, song2021scorebasedgenerativemodelingstochastic, song2020ddim,yuan2023spatio} have shown great promise in mobility trajectory modeling due to their ability to capture the randomness and uncertainty of urban mobility~\cite{zhu2023difftraj, chu2024simulating,yuan2024urbandit}. Meanwhile, their step-by-step denoising process allows for the learning of complex distributions, aligning naturally with the intricacies of human behaviors. Despite this progress, most diffusion model-based approaches~\cite{zhu2023difftraj} apply noise in a simplistic manner, often mirroring techniques from image generation where noise is independently and identically distributed (i.i.d.) across points in the trajectory. We argue that this i.i.d. noise sampling, as commonly used in models like DDPM~\cite{ho2020ddpm} and DDIM~\cite{song2020ddim}, is inadequate for urban mobility generation. As the origin of the generative process, noise plays a crucial role, significantly impacting the realism and quality of the output~\cite{qiu2024freenoisetuningfreelongervideo, tumanyan2022plugandplaydiffusionfeaturestextdriven, ge2023preserve,zhang2024trip}. As illustrated in Figure~\ref{fig:noise_map}, our analysis demonstrates a strong correlation between the initial noise and the final generated trajectory. The naive assumption of i.i.d. noise disregards the complex dependencies between trajectory points, leading to suboptimal outcomes in the generated mobility data. 

To address this, we propose leveraging more informative noise priors, which provide richer guidance to the denoising process.
To construct effective noise priors, it is essential to capture the underlying characteristics of urban mobility. However, it is not trivial due to two key challenges: (1) Movement transitions exhibit intricate spatiotemporal correlations, which cannot be easily defined by simple rules or equations, and (2) Movement behaviors in the urban environment are not independent; individual behaviors are deeply interconnected with broader social and environmental dynamics. 
Addressing these complexities is crucial for developing effective noise priors.

In this work, we propose CoDiffMob, a \textit{\uline{Diff}usion-based method for \uline{Mob}ility Generation with \uline{Co}llaborative Noise Priors}.
The core idea is to integrate both individual preferences and collective patterns into the noise prior.
Specifically, we employ a two-stage collaborative mechanism to construct noise priors. First, we utilize collective movement patterns to sample location transition sequences with a collaborative rule-based method. Then, the sequences are mapped to noise space and fused with white noise to build collaborative noise priors.
This allows the generation process to be guided more effectively by these richer, more informative noise priors that embed the interactions between individual behaviors and population dynamics. As a result, the generated urban mobility not only captures individual movement characteristics but also aligns collective flows with broader population dynamics, enabling realistic mobility generation from both individual and collective perspectives. 

We summarize our main contributions as follows:

\begin{itemize}[leftmargin=*]
    \item To the best of our knowledge, we are the first to highlight the critical role of noise sampling for urban mobility generation.
    \item We propose CoDiffMob, a novel diffusion model with collaborative noise priors that effectively incorporate spatiotemporal prior knowledge, capturing both individual movement behaviors and broader population dynamics. 
    \item Extensive experiments show the superior performance of CoDiffMob in generating urban mobility data, capturing both individual and collective patterns, with an average improvement of over 32\%. The generated data ensures privacy protection and can effectively replace real-world data to enhance downstream tasks.

\end{itemize}

\section{Preliminaries and Problem Statement}

\subsection{Urban Mobility}
Urban mobility refers to how people move within cities, typically characterized by individual trajectories and collective flows.

\textbf{Definition 1: (Individual Trajectory).}
The individual trajectory of a person is defined as a sequence of spatiotemporal points visited within a day, denoted as $\boldsymbol{x} = \{x_1, x_2, ..., x_{n}\}$,
where each point $x_i$ is defined as $x_i = (l_i, t_i)$, with $t_i$ representing the timestamp and $l_i$ representing a specific geographic coordinate.

\textbf{Definition 2: (Collective Flow).}
The collective flows of a mobility dataset $\mathcal{X}$ are defined as a matrix $\mathcal{F}^{\mathcal{X}} \in \mathbb{R}^{N\times N}$, where $N$ is the number of locations and each row of the matrix $\mathcal{F}^{\mathcal{X}}_{l_i}$ represents the population count traveling from $l_i$ to other locations, as obtained from the dataset.

\subsection{Diffusion Model for Mobility Generation}
\label{sec:pre_diff}
\label{sec:ddim}
The diffusion probabilistic model is an increasingly prominent generative framework, celebrated for its effectiveness across various generation tasks \cite{dhariwal2021diffusion, rombach2022high}. It typically comprises two core processes: the forward process, which progressively disrupts the data $\boldsymbol{x}_0$ by adding Gaussian noise:

\begin{equation}
\label{eq:diff-forward}
\begin{array}{l}
q(\bm{x}_{1:K}\,|\,\bm{x}_0) = \prod_{k=1}^{K} q(\bm{x}_k \,|\, \bm{x}_{k-1}), 
\vspace{0.5em} \\
q(\bm{x}_k \,|\, \bm{x}_{k-1}) = \mathcal{N}(\bm{x}_k;\sqrt{1-\beta_k}\bm{x}_{k-1},\beta_k\bm{I}),
\end{array}
\end{equation}

\noindent where the parameter $\beta_k \in (0,1)$ and increases as k increases, to back-propagate the gradient of the calculation, a reparameterization trick is adopted and $\bm{x}_k$ can also be given by:
\begin{equation}
\label{eq:diff-forward}
\begin{array}{l}
\boldsymbol{x}_k =  \sqrt{\alpha_k}\cdot\boldsymbol{x}_0 + \sqrt{1 - \alpha_k}\cdot \boldsymbol{z},
\vspace{0.5em} \\
\text {where }  \boldsymbol{z} \sim \mathcal{N}(\bm{0}, \bm{I}) \text{ and } \alpha_k = \prod_{i=1}^{k}(1-\beta_i).
\end{array}
\end{equation}

The denoising process gradually recovers the original trajectory from the noisy data $\bm{x}_K \sim \mathcal{N}(\bm{0}, \bm{I})$, ultimately producing trajectory data that conforms to the real distribution~\cite{ho2020ddpm, song2020ddim}:

\begin{equation}
\label{eq:diff-denoise}
\begin{array}{l}
\boldsymbol{x}_{k-1} =  \sqrt{\alpha_{k-1}}(\frac{\boldsymbol{x}_k - \sqrt{1-\alpha_k}\epsilon_{\theta}^{(k)}(\boldsymbol{x}_k)}{\sqrt{\alpha_k}}) 
\vspace{0.4em} \\
\quad \quad \quad \text{} + \sqrt{1 - \alpha_{k-1} - \sigma_k^2}\cdot \epsilon_{\theta}^{(k)}(\boldsymbol{x}_k) + \sigma_k\epsilon_k,
\vspace{0.3em} \\
\text {where }  \epsilon_k \sim \mathcal{N}(\bm{0}, \bm{I}) \text{ and } \sigma_k = \sqrt{\frac{(1-\alpha_k / \alpha_{k-1})(1-\alpha_{k-1})}{1-\alpha_{k}}},
\end{array}
\end{equation}

\noindent here $\epsilon_{\theta}$ is a trained neural network to predict the noise added to the real data. 
By repeating this process, we can derive the approximate true data $\bm{x}_0$ from the Gaussian noise $\bm{x}_K$.

Given the real-world trajectory dataset $\mathcal{X} = \{\boldsymbol{x}^1, \boldsymbol{x}^2, ..., \boldsymbol{x}^{n} \}$, the objective of mobility generation task is to learn a generative model $\bm{G}_{\theta}$ that can generate synthetic mobility trajectories $\mathcal{Y} = \{\boldsymbol{y}^1, \boldsymbol{y}^2, ..., \boldsymbol{y}^{n} \}$. By training a denoiser $\epsilon_{\theta}$ based on dataset $\mathcal{X}$, synthetic trajectories can be sampled by the denoising process shown in Equation~\ref{eq:diff-denoise}, we introduce the DDIM~\cite{song2020ddim} sampling method by setting the noise parameter $\sigma_k$ to $0$ to speedup the sampling process:

\begin{equation}
\label{eq:ddim}
\begin{array}{r}
\bm{y} =  \text{DDIM}_{\theta}(\bm{z}) \quad \text { where }  \bm{z} \sim \mathcal{N}(\bm{0}, \bm{I}).
\end{array}
\end{equation}

Under this setting, the stochastic nature of the denoising generation process is eliminated, and the noise and the generated trajectory are one-to-one correspondence.

\section{Analysis and Motivation}
\label{sec:analysis}
In this section, we analyze the relationship between the noise and the trajectory data generated from it by the diffusion model. As illustrated in section ~\ref{sec:pre_diff}, in DDIM, each generated trajectory corresponds to a unique noise. Here, we obtain the noise corresponding to the real-world trajectories by reversing the denoising process shown in equation~\ref{eq:diff-denoise}.
We conduct an in-depth analysis of the correlation between trajectory data $\bm{x}_0$ and noise $\bm{x}_K$, the results are shown below:

\subsection{From Noise to Movement Semantics}
To directly observe the relationship between the characteristics of trajectories and the location of their corresponding noise in noise space,
we perform t-SNE \cite{van2008visualizing} on this noise to facilitate visualization. 
Figure \ref{fig:noise2traj} shows the visualization results. The blue dots in the background represent randomly sampled data points from an i.i.d Gaussian distribution. The dots with other different colors denote the noise obtained from real trajectories with varying movement patterns. 

We find that the noise location in the noise space differs based on the movement patterns of the trajectory: the three red dots on the right correspond to three trajectories from the same person, whose movement radius is within 2 km with only three distinct locations. The two green dots at the bottom left represent two trajectories with similar movement patterns, where people travel horizontally across the entire city and return within a day. The orange dots at the top point to another travel habit, where the person makes extensive movements throughout the day, visiting multiple areas across the city. 
These results indicate a strong correlation between the initial noise and the final generated trajectory data. 

\subsection{From Noise to Spatio-Temporal Patterns}

\subsubsection{\textbf{Spatial Domain}}

For a single real trajectory $\bm{x}_0$ and the corresponding noise $\bm{x}_K$, the sequence of its visited locations $\{l_1, l_2, ..., l_{n}\}$ corresponds to a series of points $\{p_1, p_2, ..., p_{n}\}$ in the noise space. For two consecutive locations in time, if $l_{i-1} \neq l_{i}$, we consider that the individual has moved with a 2D vector $\Vec{m} = l_{i} - l_{i-1}$. In this case, we record the \textit{moving direction} $\arctan{(m_y,m_x)}$ and \textit{travel distance} $\lVert \Vec{m} \rVert$ of this movement in the real world, as well as the relative position information between the points $p_{i-1}$ and $p_i$ in the noise space. We conduct a linear regression analysis on the recorded movement features in the real world and their corresponding relative positions in the noise space. The regression results are shown in Figure \ref{fig:corr_spatial}.

The results indicate a strong correlation between the real-world moving direction and the relative direction of points in noise space, with $R^2 = 0.9552$. This suggests that in the generated trajectory, the direction of each movement is largely determined by the corresponding direction in the initial noise sequence. This may be because the distribution of moving directions in the real trajectory dataset is relatively uniform, leading the generation model to forego encoding the distribution of moving directions and instead directly correlate them with the moving directions in the initial noise sequence.

For travel distance, $R^2 = 0.2813$, which indicates a correlation coefficient of more than $0.5$, showing that there is a certain correlation between the distances in real trajectory points and the distances between corresponding points in the noise space, although not as strong as in moving direction. This is because the travel distances in the real data largely follow a specific power-law distribution, which the generative model captures and uses to generate trajectory data. 

\begin{figure*}[t]
\begin{center}
\includegraphics[width=0.9\linewidth]{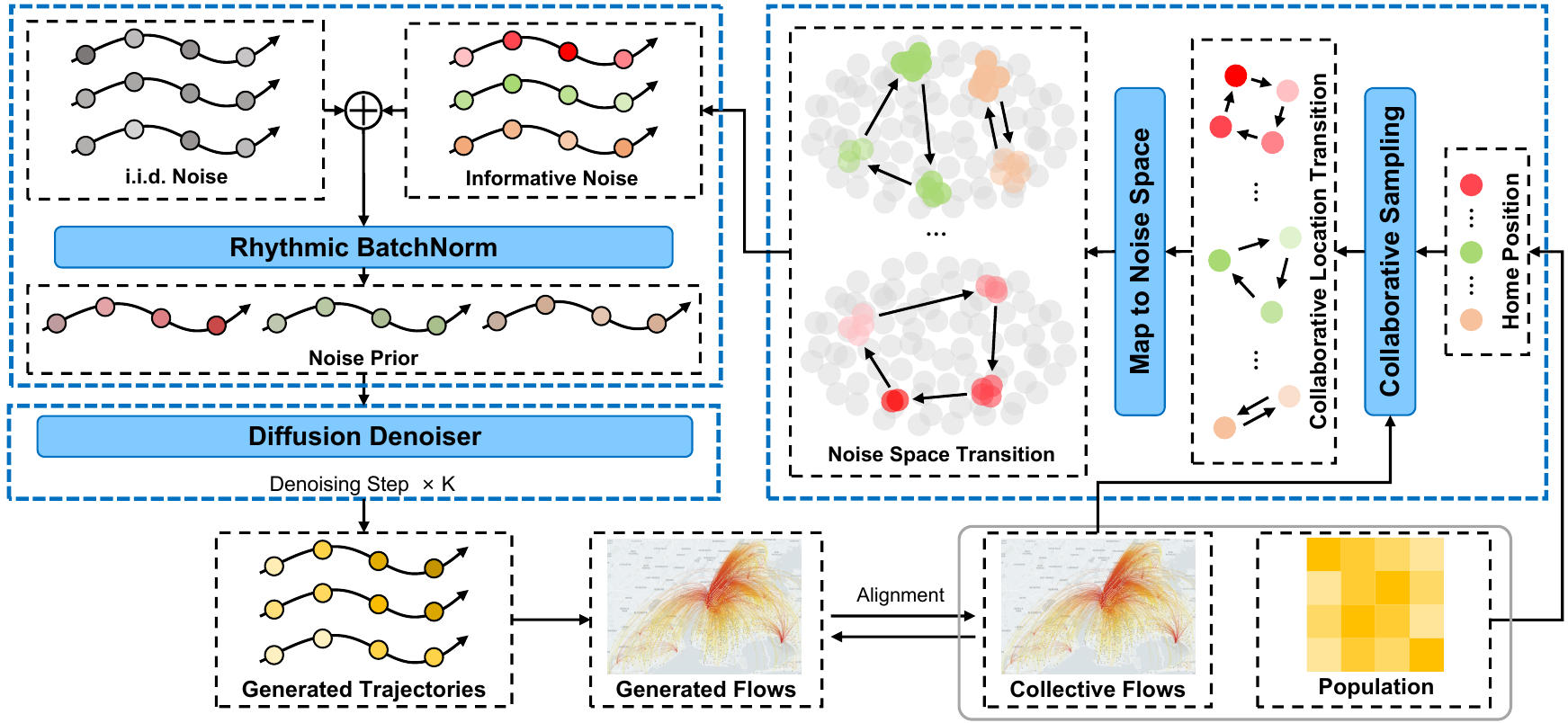}
\end{center}
\caption{The overall workflow of mobility generation with collaborative noise priors.}
\label{fig:model}
\end{figure*}

\subsubsection{\textbf{Temporal Domain}}
We analyze the relationship between the probability of people in the entire trajectory dataset $\mathcal{X}$ engaging in movement at specific times and the overall variance of the corresponding noise sequences at those times.
The moving probability of users at time $t$ is defined as: 

\begin{equation}
\label{eq:move-prob}
P_{move}^{t} = \frac{\sum_{\boldsymbol{x} \in \chi} \sum_{t_i=t} \mathbb{I}(l_i \neq l_{i+1})}{|\mathcal{X}|}.
\end{equation}

Figure \ref{fig:corr_time} presents the visualization of the relationship between variance and moving probability. The horizontal axis represents different times of the day. The yellow line shows the overall moving probability of users at different times, while the blue line represents the overall noise variance at corresponding times. 
The overall moving probability of people largely reflects the temporal patterns of collective mobility within the city: before dawn, most people remain stationary at home, and after sunrise, various travel activities commence, with three distinct peaks in the morning, midday, and evening. After nightfall, mobility settles back into calm. 
We surprisingly find that the noise variance follows this same rhythmic pattern, the magnitude of the noise variance changes following the variations in crowd movement probabilities.

\section{Method}
In this section, we elaborate on the proposed urban mobility generation method, CoDiffMob. 
We train a diffusion model to capture individual movement characteristics and employ a two-stage collaborative mechanism to construct noise priors that incorporate collective movement information, thereby bridging the influence of population dynamics and individual behaviors.

\subsection{The Overall Workflow}
\label{sec: individual_gen}

The overall workflow of CoDiffMob is shown in Figure~\ref{fig:model}, our model generates urban mobility  through a diffusion denoising process with collaborative noise priors:

\begin{equation}
\label{eq:ddim-np}
\begin{array}{l}
\bm{z}_P =  \text{Collab}_{n}(\text{Collab}_{\pi}(\mathcal{F},\mathcal{P}),\,\bm{z}),
\\
\bm{x}_0 =  \text{DDIM}_{\theta}(\bm{z}_P) \quad \text{where } \bm{z} \sim \mathcal{N}(\bm{0}, \bm{I}),
\end{array}
\end{equation}

\noindent where the noise prior $\bm{z}_P$ is constructed through a two-stage collaborative mechanism, \textit{collaborative transition sampling} through policy $\pi$, 
and \textit{collaborative noise fusion} with i.i.d noise $\bm{z_{i.i.d.}}$.

Following ~\cite{zhu2023difftraj}, we apply a UNet-based diffusion denoiser to generate individual trajectory from $\bm{z}_p$.
During the network's computation, the noised trajectory is represented as a $T \times C$ vector, where $T$ is the length of the trajectory, and $C$ is the dimensionality of the location information at each moment (for latitude and longitude, $C=2$). During the upsampling and downsampling stages of the UNet, the model leverages one-dimensional convolutional neural networks to model the spatiotemporal correlations between trajectory points. Additionally, an attention mechanism is applied to the features after downsampling to capture relationships between the data in the latent space.

\subsection{Collaborative Noise Priors}
To accurately model urban mobility, a collaborative approach that integrates both individual behavior and collective movement is essential. We focus on the noise in the diffusion denoising process and introduce a two-stage collaborative mechanism to construct noise priors for mobility generation.

As shown in Figure~\ref{fig:model}, a rule-based collaborative strategy is first employed to sample individual location transition sequences informed by the patterns of collective flows. Subsequently, the transition sequences are mapped to the noise space, resulting in informative noise that encapsulates collective movement patterns. This noise is then fused with i.i.d noise to produce the final noise prior. The detailed process is as follows:

\subsubsection{\textbf{Collaborative Transition Sampling}}
The diffusion model generates urban mobility in the form of individual trajectories, and the noise priors must be constructed in a similar form. In this process, we first need to convert the collective movement information represented by the flows $\mathcal{F}$ into individual location transition sequences.
The exploration and preferential return (EPR) model~\cite{song2010modelling, jiang2016timegeo} is a widely used rule-based model to simulate individual movement. Building upon this, we design an individual-collective collaborative simulation approach, integrating flow information into the sampling process of individual location transitions.

In the EPR model, each individual's actions fall into four categories: \textit{stay}, \textit{home return}, \textit{preferential return}, and \textit{explore}, representing four distinct behaviors: staying in the same location, returning home, revisiting previously visited locations, and exploring new places. The original EPR model chooses new places based on the distribution of travel distance~\cite{jiang2016timegeo}, which makes it hard to accurately describe the collective movement patterns. We design a collaborative model to combine individual and collective patterns. During the sampling process, the model first determines an individual's home location $l_1$ based on the population $\mathcal{P}$ and then simulates the individual's behavior using a collaborative strategy as follows:

\begin{equation}
\label{eq:ddim-np}
\vspace{-0.3em}
\begin{array}{r}
\pi(l_i \,|\, l_{<i},t_i) =  \text{Collab}(\pi_I(l_i\,|\,l_{<i},t_i),\pi_{F}(l_i\,|\,\widetilde{\mathcal{F}}_{l_{i-1}})).
\end{array}
\end{equation}

The collaborative strategy combines the individual behavior of preferential return and the exploration with collective movement patterns.
When an individual chooses to explore new locations, the strategy $\pi$ follows the flow-based policy $\pi_{F}(l_i\,|\,\widetilde{\mathcal{F}}_{l_{i-1}})$ and select target location based on the distribution of collective movements $\widetilde{\mathcal{F}}_{l_{i-1}}$, which is computed from flow $\mathcal{F}_{l_{i-1}}$. Otherwise, the strategy follows the individual policy $\pi_I(l_i\,|\,l_{<i},t_i)$ given by the original EPR model and chooses the next location based on historical preference.


\subsubsection{\textbf{Collaborative Noise Fusion}}
Currently, the majority of diffusion models for mobility generation tasks are trained on individual trajectories and use i.i.d. Gaussian noise during the sampling process. While this approach performs good in generating individual trajectories, it fails to capture the patterns of collective movement. The sampling method based on EPR and flow $\mathcal{F}$, as introduced earlier, can incorporate collective information into individual location transitions $\boldsymbol{x}_{\mathcal{F}}\,$. However, rule-based models often fail to model individual behavior effectively. To address this, we fuse both of the information within the noise space to construct noise priors for mobility generation with the diffusion model.

As indicated in section~\ref{sec:analysis}, by applying the reverse DDIM process on the location transition sequence, we can obtain the informative noise $\bm{z}_{\mathcal{F}}$ corresponding to $\boldsymbol{x}_{\mathcal{F}}\,$.
Direct usage of the noise $\bm{z}_{\mathcal{F}}$ itself leads to poor performance on modeling individual behaviors.
To achieve collaborative modeling of both individual behavior and collective movement,
we fuse $\bm{z}_{\mathcal{F}}\,$ with Gaussian noise $\bm{z_{i.i.d.}}$ to obtain the collaborative noise prior $\bm{z}_P$ for mobility generation:

\vspace{-0.5em}
\begin{equation}
\label{eq:noise_fusion}
\vspace{-0.5em}
\begin{array}{l}
\bm{z_{i.i.d.}} \sim \mathcal{N}(\bm{0}, \bm{I}), \\
\bm{z_{\mathcal{F}}} = \text{Inverse-DDIM}_{\theta}(\boldsymbol{x}_{\mathcal{F}}), \\
\bm{z}_p =  \bm{z_{i.i.d.}} + \bm{z_{\mathcal{F}}}.
\end{array}
\end{equation}

\subsubsection{\textbf{Rhythmic Noise BatchNorm}}
In addition to incorporating collective transition patterns in the spatial domain, we also model the rhythmic information of crowd activity into the noise prior by controlling the noise variance. During the trajectory generation process, we organize the input data in batches.
We apply batch normalization based on the rhythmic information to this data:

\vspace{-0.5em}
\begin{equation}
\label{eq:noise_norm}
\begin{array}{l}
\mu_t = \frac{1}{B}\sum_{i=1}^{B}\bm{z}_p^{i,t},\quad \sigma_t = \sqrt{\frac{1}{B}\sum_{i=1}^{B}(\bm{z}_p^{i,t} - \mu_t)^2},
\vspace{0.2em} \\
\bm{z}_p^t = \frac{\bm{z}_p^t - \mu_t}{\sigma_t} \times \mathcal{R}_t \quad \text{where } \mathcal{R}_t \propto P_{move}^t \text{ and } t = 1, \ldots ,T.
\end{array}
\end{equation}

The variance of the noise at each time step $t$ is normalized to be proportional to the crowd moving probability calculated by Equation~\ref{eq:move-prob} at the corresponding time, as shown in Figure~\ref{fig:corr_time}.

\subsection{Training and Inference}
During the training process of the diffusion model, noise based on i.i.d. Gaussian distribution is added to the input trajectories using a uniformly sampled noise level $k$. The model takes the noisy trajectory and conditions that include the noise level and the starting point of the trajectory as inputs. Based on this information, the model predicts the noise added to the trajectory, enabling the gradual denoising process to restore the individual trajectory from pure noise, details can be found in Appendix~\ref{appendix:train}.
The inference process is divided into two stages, the collaborative noise prior sampling and the individual trajectory generation. Details are shown in Algorithm~\ref{alg:generating}:

\begin{algorithm}[ht]
\caption{Mobility Generation with Collaborative Noise Priors}
\label{alg:generating}
    \begin{algorithmic}[1]
        \STATE \textbf{Require} EPR model parameters $\{n_{\omega}, \beta_1, \beta_2, P, \rho, \gamma\}$, collective flows $\mathcal{F}$, population distribution $\mathcal{P}$, trajectory time interval $t$, number of trajectory points $n$, collaborative sampling strategy $\pi$, trajectory length $T$, diffusion denoiser $\epsilon_{\boldsymbol{\theta}}$, denoising step $K$
        \\
        \STATE \textbf{Collaborative Noise Priors Sampling}
        \STATE Initialize home point $(l_1,t_1)$ based on $\mathcal{P}$
        \FOR{$i = 2, \ldots, n$}
            \STATE $t_i = t_{i-1} + t$
            \STATE $l_{i} = \pi(l_{<i},t_i\,|\,n_{\omega}, \beta_1, \beta_2, P, \rho, \gamma, \mathcal{F})$
        \ENDFOR
        \STATE $\boldsymbol{x}_{\mathcal{F}} = \{(l_1,t_1), (l_2,t_2), ..., (l_n,t_n)\}$
        \STATE \text{informative noise } $\boldsymbol{z}_{\mathcal{F}} = $ Inverse-DDIM $(\boldsymbol{x}_{\mathcal{F}})$
        \STATE \text{noise prior } $\bm{z}_p = \text{\textit{Rhythmic BN}}(\bm{z_{i.i.d.}} + \boldsymbol{z}_{\mathcal{F}})$
        \STATE \textbf{Individual Trajectory Generation}
        \STATE Initialize $\bm{x}_K = \bm{z}_p$
        \FOR{$k = K, \ldots, 1$}
            \STATE $\epsilon_k = \epsilon_{\theta}(\bm{x}^k, k \,|\, l_1)$
            \STATE $\boldsymbol{x}_{k-1} =  \sqrt{\alpha_{k-1}}(\frac{\boldsymbol{x}_k - \sqrt{1-\alpha_k}\epsilon_k} {\sqrt{\alpha_k}}) + \sqrt{1 - \alpha_{k-1}- \sigma_k^2}\cdot \epsilon_k$
        \ENDFOR
        \STATE Execute trajectory sequence using output $\bm{x}_0$
    \end{algorithmic}
\end{algorithm}
\vspace{-1em}

\begin{table*}[ht]
\footnotesize
\centering
\caption{Performance comparison of different generation models, where \textbf{bold} denotes best results and \underline{underline} denotes the second best results.}
\vspace{-3mm}
\resizebox{1.0\textwidth}{!}{
\begin{tabularx}{\textwidth}{l|YYYYYYY|YYYYYYY}
\toprule
\textbf{Dataset} & \multicolumn{7}{c}{\textbf{ISP}} & \multicolumn{7}{c}{\textbf{MME}} \\
\midrule
\multirow{2}{*}[-0.8ex]{\textbf{Metrics}} & \multicolumn{4}{c}{Trajectory} & \multicolumn{2}{c}{Flow} & \multirow{2}{*}[-0.8ex]{Diversity} & \multicolumn{4}{c}{Trajectory} & \multicolumn{2}{c}{Flow} & \multirow{2}{*}[-0.8ex]{Diversity} \\ 
\cmidrule(l){2-5} \cmidrule(l){6-7} \cmidrule(l){9-12} \cmidrule(l){13-14} 
& Radius & Distance & Duration & DailyLoc & CPC    & MAPE & & Radius & Distance & Duration & DailyLoc & CPC    & MAPE  \\ 
\midrule
TimeGEO      & 0.3171 & 0.2984 & \underline{0.0526} & 0.2877 & 0.2133 & 0.8637 & 0.1498
             & 0.3678 & 0.4807 & \underline{0.0891} & \underline{0.2512} & 0.3248 & 0.8274 & 0.1682 \\ 
PateGail     & 0.1979 & 0.1826 & 0.1867 & 0.2933 & 0.0714 & 0.9827 & 0.0984  
             & 0.1824 & 0.1673 & 0.1376 & 0.2883 & 0.1548 & 0.9136 & 0.1136 \\ 
MoveSim      & 0.2313 & 0.2623 & 0.2861 & 0.4289 & 0.0932 & 0.9983 & 0.0492 
             & 0.1936 & 0.2219 & 0.2084 & 0.3746 & 0.1739 & 0.8873 & 0.0492 \\
VOLUNTEER    & 0.4435 & 0.4716 & 0.4245 & 0.5155 & 0.0044 & 1.0088 & 0.0839 
             & 0.3826 & 0.3911 & 0.3512 & 0.4558 & 0.0927 & 0.9512 & 0.0928 \\
TrajGDM      & 0.2977 & 0.2393 & 0.2443 & 0.4121 & 0.0070 & 1.0022 & 0.1271 
             & 0.3123 & 0.2577 & 0.2875 & 0.3961 & 0.0526 & 0.9813 & 0.1474 \\
DiffTraj     & \underline{0.0519} & \underline{0.1662} & 0.0895 & \underline{0.1617} & \underline{0.3992} & \underline{0.8331} & \textbf{0.0028} 
& \underline{0.1028} & \underline{0.1242} & 0.0949 & 0.2657 & \underline{0.4198} & \underline{0.7727} & \textbf{0.0053} \\
\midrule
CoDiffMob    & \textbf{0.0183} & \textbf{0.1203} & \textbf{0.0245} & \textbf{0.1558} & \textbf{0.6842} & \textbf{0.6361} & \underline{0.0046}
& \textbf{0.0519} & \textbf{0.0970} & \textbf{0.0796} & \textbf{0.1729} & \textbf{0.6067} & \textbf{0.6731} & \underline{0.0089} \\
\bottomrule
\end{tabularx}
\label{table:model_perf}
}
\end{table*}

\begin{figure*}[t]
\begin{center}
\includegraphics[width=0.85\linewidth]{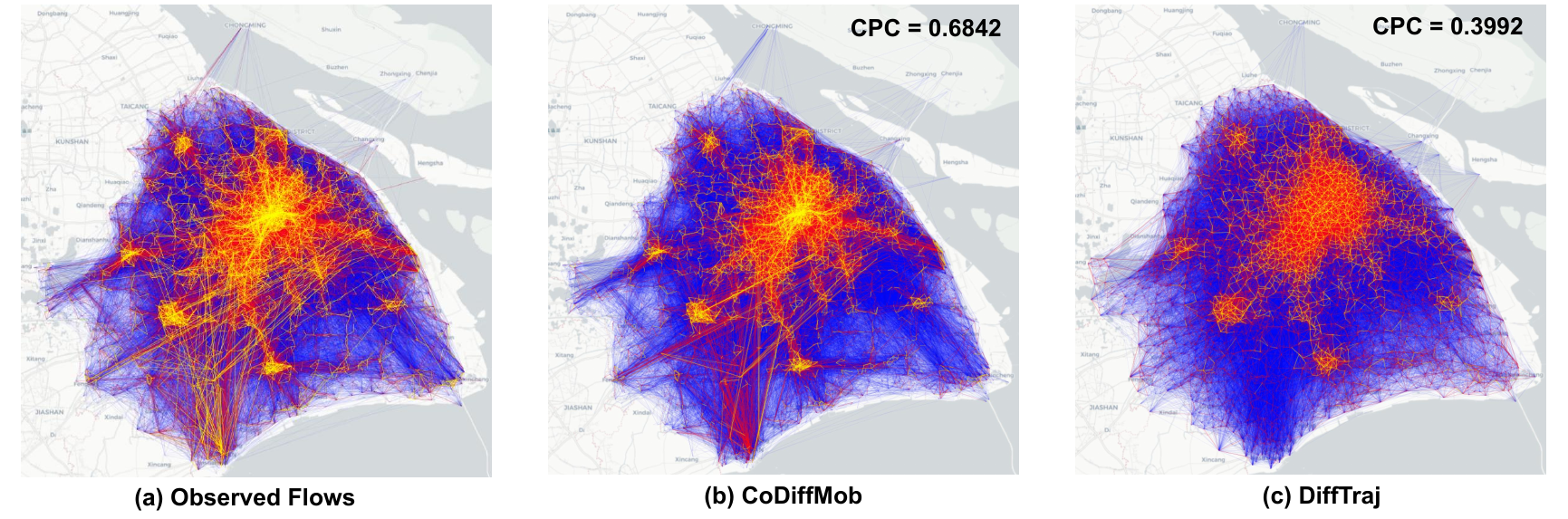}
\end{center}
\vspace{-3mm}
\caption{Visualization of the observed flows and generated flows on ISP dataset.}
\label{fig:flow_vis_ISP}
\end{figure*}

\section{Experiments}
We conduct comprehensive experiments to validate the effectiveness of  CoDiffMob to answer the following four research questions:

\begin{itemize}[leftmargin=*]
    \item \textbf{RQ1}: Can  CoDiffMob generate realistic urban mobility data that captures both individual behaviors and collective movement patterns?
    \item \textbf{RQ2}: Are there risks of exposing users' privacy in the generated mobility data?
    \item \textbf{RQ3}: Can the generated mobility data effectively support downstream tasks in the same way as real-world data?
    \item \textbf{RQ4}: How do different parts of our proposed method contribute to the final performance?
\end{itemize}

\subsection{Experimental Setup}

\subsubsection{\textbf{Datasets.}}
We use two real-world mobility datasets (\textit{ISP} and \textit{MME}) to evaluate the performance of CoDiffMob, which are collected from Shanghai and Nanchang through web-based applications, the details of the datasets can be found in Appendix~\ref{appendix:dataset}.

\subsubsection{\textbf{Metrics.}}
We use metrics from two aspects: For individual trajectories, we evaluate the generated trajectories in terms of \textit{Distance}, \textit{Radius}, \textit{Duration}, and \textit{DailyLoc}~\cite{feng2020learning,li2023learning}.
For generated flows, we calculate \textit{Common Part of Commuters} (CPC)~\cite{simini2021deep} and \textit{MAPE} between the real and generated population flows. 
Furthermore, we introduce a \textit{Diversity} metric, to validate the diversity of the model's generated results.
Details of the calculation can be found in Appendix~\ref{appendix:metrics}.

\subsubsection{\textbf{Baselines.}}
We compare the performance of CoDiffMob with six state-of-the-art baselines for mobility generation including \textbf{TimeGEO} \cite{jiang2016timegeo}, \textbf{PateGail} \cite{wang2023pategail}, \textbf{MoveSim} \cite{feng2020learning}, \textbf{VOLUNTEER} \cite{long2023practical},\textbf{TrajGDM} \cite{chu2024simulating}, and \textbf{DiffTraj} \cite{zhu2023difftraj}.
Details about these methods can be found in Appendix~\ref{appendix:baselines}.

\subsection{Overall Performance (RQ1)}
The overall performance of CoDiffMob compared to baselines is shown in Table \ref{table:model_perf}. Based on the experimental results, we draw the following conclusions:

\begin{itemize}[leftmargin=*]
\item \textbf{CoDiffMob outperforms baselines in mobility trajectory generation.} On the spatial scale, data generated by our model can more closely match the distribution of travel distances and radius with the two real-world mobility datasets, achieving an average improvement of over 40\%. On the temporal scale, our model can more effectively capture daily activity patterns, the performance in terms of location duration time and the number of locations visited per day both outperform the best baseline method with an average improvement of over 31\%.
\item \textbf{CoDiffMob achieves significant improvements over baselines in reproducing collective patterns.} Baseline methods struggle to effectively capture the population dynamics. Whether evaluating flow similarity or relative error in transition probabilities, the collective features generated by baseline methods show considerable deviation from real data, The CPC metric is generally below $0.4$, while the MAPE metric is predominantly above $0.8$. Our model can successfully capture collective movements through the collaborative noise prior, achieving CPC greater than $0.6$ and MAPE less than $0.7$ on both datasets, the average improvement is more than 35\%.
\item \textbf{The impact of introducing prior knowledge on the diversity of generated results remains within an acceptable range.} While incorporating collaborative noise priors reduces the diversity of the generated outputs, our model's performance on diversity metrics is only slightly behind DiffTraj. Additionally, it still significantly outperforms other autoregressive generation methods in terms of diversity.
\end{itemize}

Figure \ref{fig:flow_vis_ISP} shows the visualization of collective flows from the real data and generated data. We can observe that the flows generated by our method are more similar to the real ones. The image on the right shows the result generated by the best baseline, Difftraj. As can be seen, without the noise priors providing flow information, the model is only able to learn individual movement patterns.

\subsection{User Privacy Protection (RQ2)}
Directly utilizing user data collected from web-based applications and services often leads to serious privacy issues and our approach contributes to protecting the privacy of web users through data generation. To verify that the data generated by our method does not compromise user privacy, we perform experiments from different aspects including \textit{Uniqueness Testing} and \textit{Membership Inference Attacks} on the generated mobility data:

\begin{figure}[t]
    \centering
    \hspace{-1em}
    \begin{subfigure}[b]{0.5\linewidth}
        \includegraphics[width=1.0\linewidth]{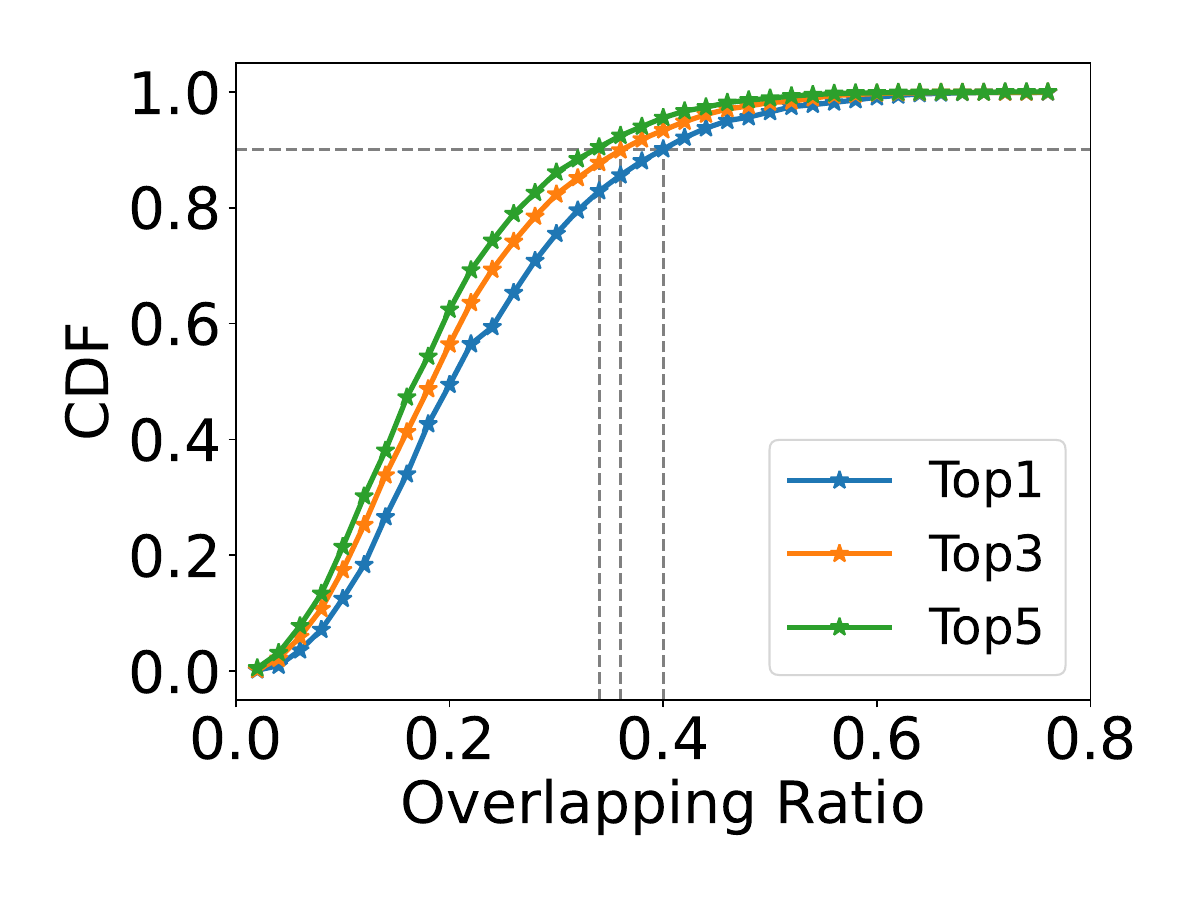}
        \vspace{-5mm}
        \caption{ISP Dataset}
        \label{fig:overlap_isp}
    \end{subfigure}
    \begin{subfigure}[b]{0.5\linewidth}
        \includegraphics[width=1.0\linewidth]{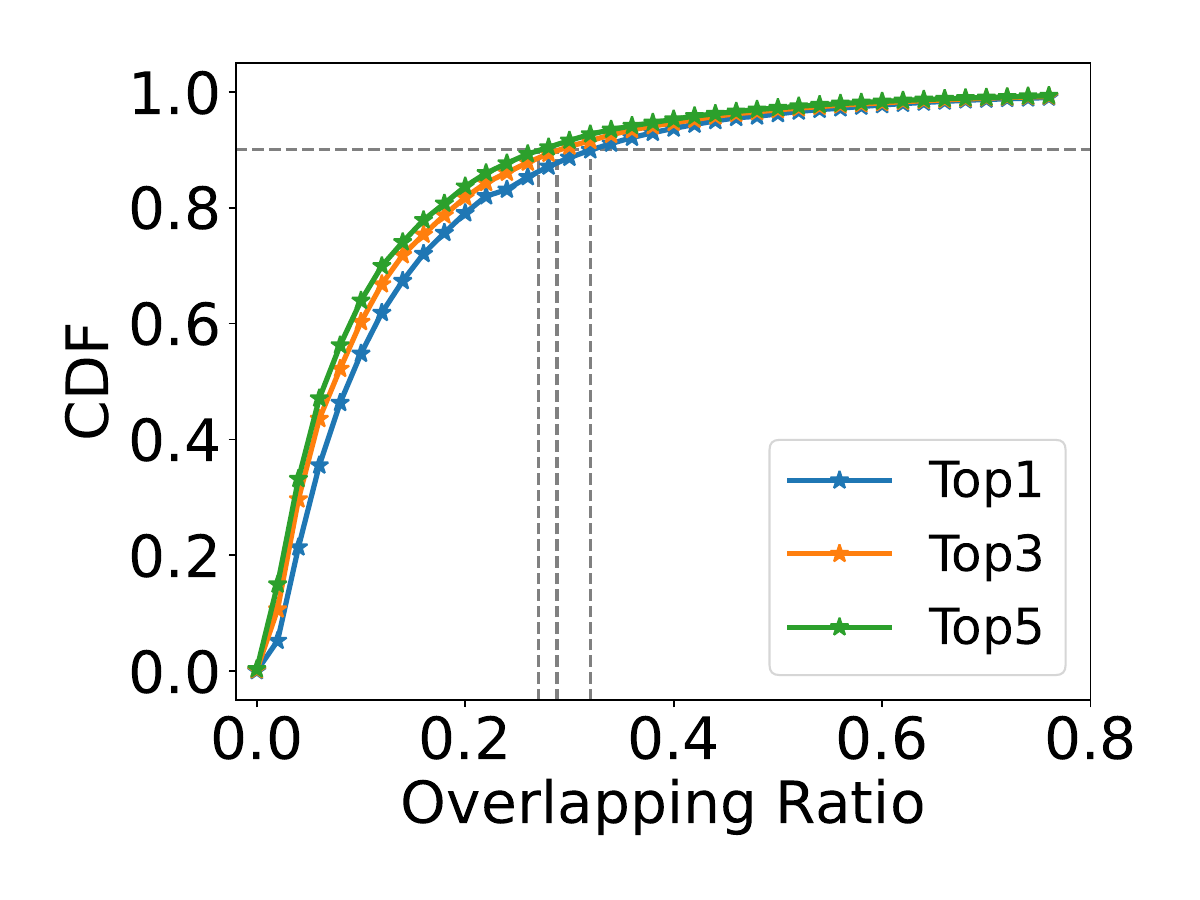}
        \vspace{-5mm}
        \caption{MME Dataset}
        \label{fig:overlap_mme}
    \end{subfigure}
    \hspace{-1em}
    \vspace{-3mm}
    \caption{Evaluation results of uniqueness testing.}
    \label{fig:overlap}
\end{figure}

\subsubsection{\textbf{Uniqueness Testing}}
We select trajectories from the generated data by random and compare them with real trajectories from the training dataset. The overlapping ratio is defined as the proportion of identical locations relative to the overall trajectory length. After that, we identify the real trajectory that is the most similar to the generated one. We then calculate the distribution of the overlapping ratio and select the top 1, 3, and 5 real trajectories that exhibit the greatest similarity to each generated trajectory. Figure~\ref{fig:overlap} shows the empirical cumulative distribution. On both datasets, more than 90\% of the generated trajectories fail to find any real trajectories with an overlapping ratio exceeding 40\%, which demonstrates that our method learns to generate entirely novel trajectories, but not merely replicating existing ones.

\subsubsection{\textbf{Membership Inference Attacks}}
\begin{figure}[t]
\begin{center}
\includegraphics[width=0.7\linewidth]{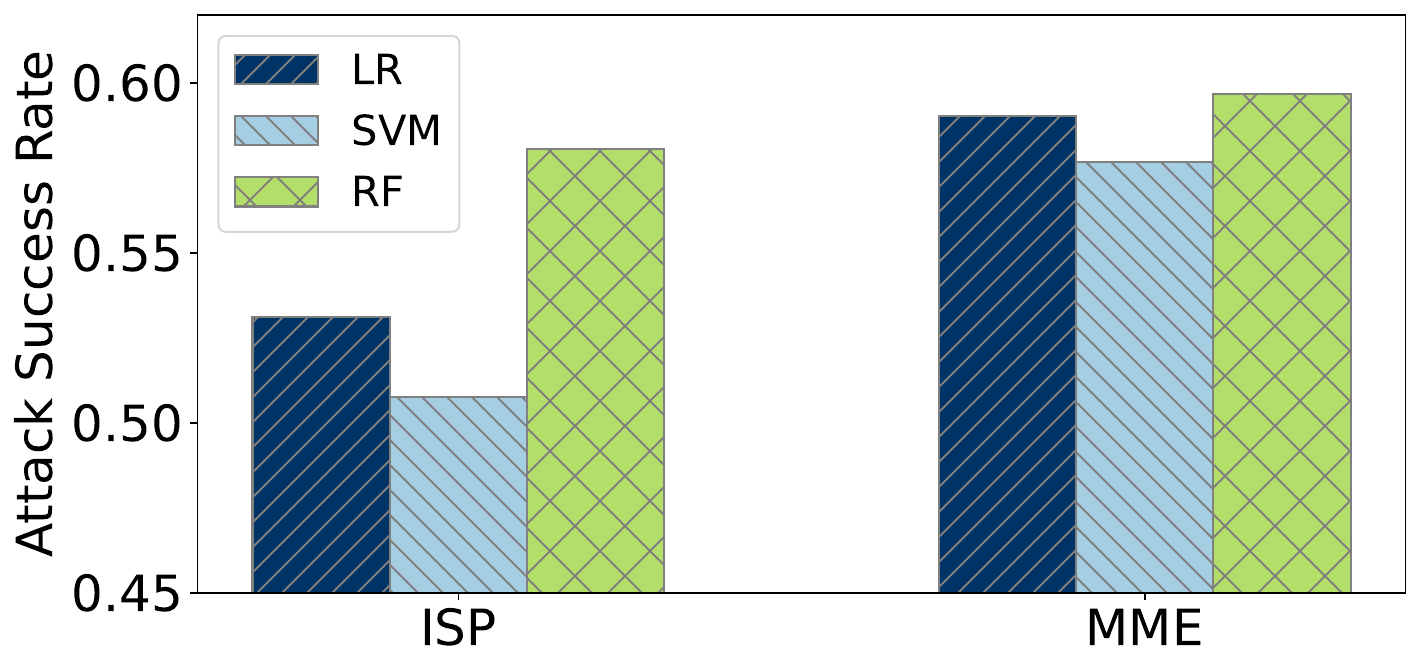}
\end{center}
\vspace{-3mm}
\caption{Success ratio of membership inference attack using different classification algorithms. }
\label{fig:mia}
\end{figure}

This attack aims to determine whether the given samples are part of the training dataset. We follow the experimental setup outlined in~\cite{Lin_2020} and~\cite{mia_2017}. To mitigate the impact of classification methods, we employ three widely used classification algorithms: Random Forest (RF), Support Vector Machine (SVM), and Logistic Regression (LR) to conduct the attack. Positive samples correspond to trajectories in the training data, and negative samples do not. The metric used to evaluate the attack is the success rate, which represents the percentage of successful trials in correctly identifying whether a sample belongs to the training dataset~\cite{Lin_2020, mia_2017}. Figure~\ref{fig:mia} presents the results of the attack. The attack success rate is consistently below $0.6$ on both datasets, indicating that the attacker struggles to determine whether trajectories are present in the training data based on the generated data. These findings highlight the robustness of our framework against membership inference attacks.

\subsection{Utility of the Generated Data (RQ3)}
Without compromising user privacy, CoDiffMob can better model urban mobility patterns, and the generated mobility data provide stronger support for downstream tasks~\cite{yuan2024unist,yi2024get,zhou2024coms2t}.
We conduct experiments on the common downstream task of mobility prediction using synthetic data. We employ both LSTM-based \cite{fattore2020automec} and Transformer-based \cite{vaswani2017attention} models for mobility prediction. The training set consists of a mixture of synthetic and real data, while the test set comprises a separate set of real trajectory data. We set different training set mix ratios and compare the downstream task performance under three settings: using only real data, using real data combined with the best baseline-generated data, and using real data combined with data generated by our model.

\begin{figure}[t]
    \centering
    \hspace{-1em}
    \begin{subfigure}[b]{0.5\linewidth}
        \includegraphics[width=1.0\linewidth]{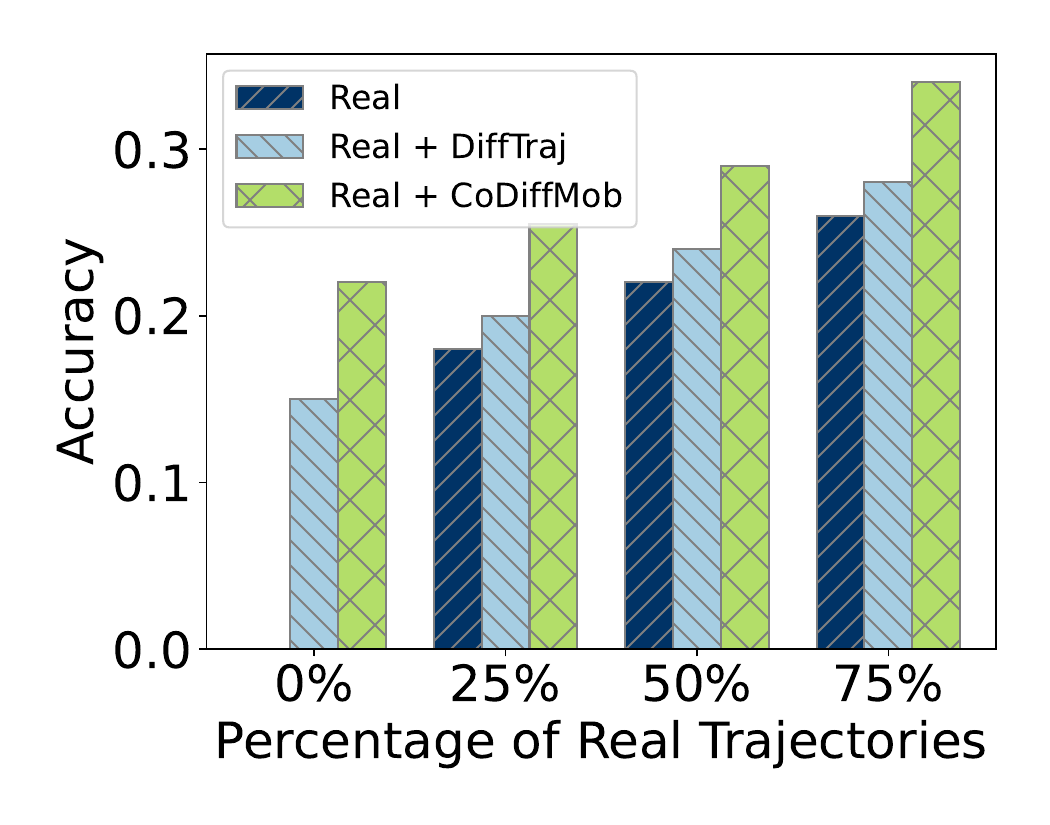}
        \caption{Transformer}
        \label{fig:utility_trans}
    \end{subfigure}
    \begin{subfigure}[b]{0.5\linewidth}
        \includegraphics[width=1.0\linewidth]{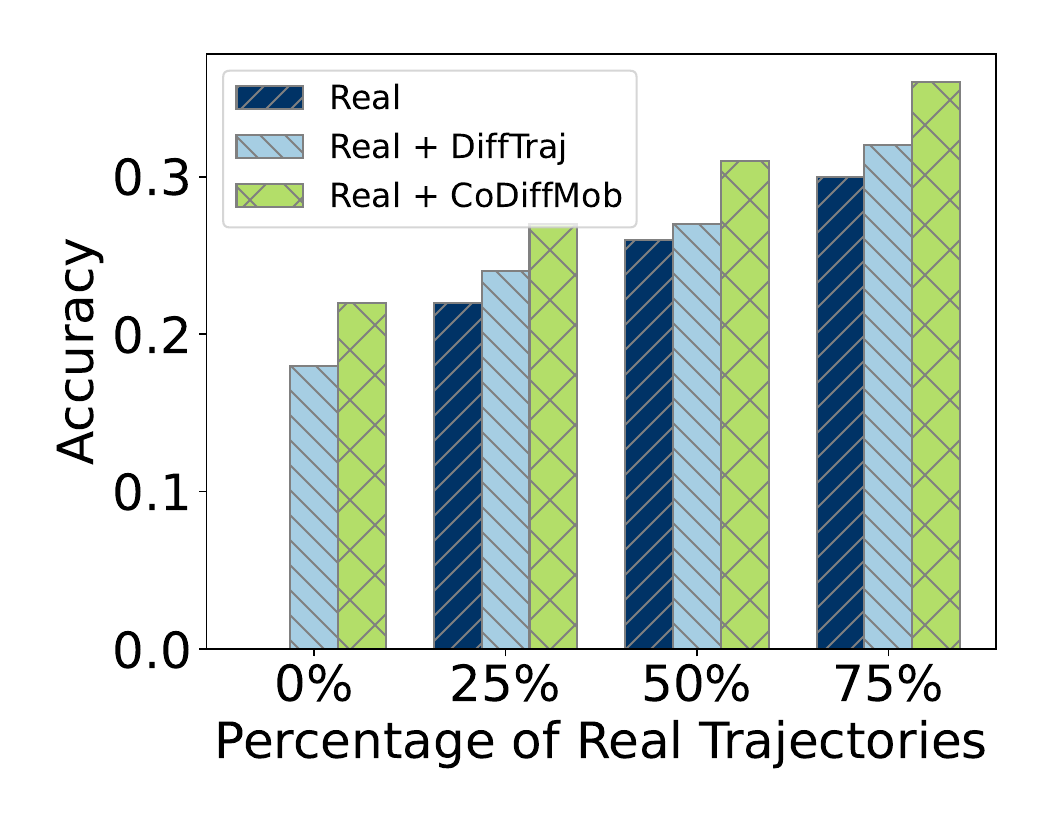}
        \caption{LSTM}
        \label{fig:utility_lstm}
    \end{subfigure}
    \hspace{-1em}
    \vspace{-3mm}
    \caption{Mobility prediction based on generated trajectories.}
    \label{fig:utility}
\end{figure}

Figure \ref{fig:utility} presents the experimental results based on the two models. The horizontal axis represents the proportion of real-world trajectory data in the training set, while the vertical axis shows the accuracy of mobility prediction. As can be seen, the inclusion of synthetic data improves downstream task performance across different data mix ratios. Additionally, compared to the data generated by the best baseline, DiffTraj, the data generated by CoDiffMob leads to more significant improvements in downstream task performance, indicating that our model generates trajectory data with characteristics more closely resembling those of the real dataset.

\begin{table}[t]
\centering
\caption{Results of the ablation study on ISP dataset.}
\vspace{-1em}
\resizebox{1.0\linewidth}{!}{
    \begin{tabular}{lcccccc}
    \toprule
    \multirow{2}{*}{} & \multicolumn{4}{c}{Trajectory} & \multicolumn{2}{c}{Flow} \\ 
    \cmidrule(l){2-5} 
    \cmidrule(l){6-7}
    & Radius & Distance & Duration & DailyLoc & CPC    & MAPE    \\ 
    \midrule
    w/o Noise Prior      & \underline{0.0673} & 0.1661 & \underline{0.0461} & \underline{0.1635} & 0.3416  & 0.8489  \\
    w/o Noise Fusion    & 0.2843 & \underline{0.1616} & 0.0561 & 0.4433 & \textbf{0.7165}  & \textbf{0.5351}  \\
    CoDiffMob          & \textbf{0.0183} & \textbf{0.1203} & \textbf{0.0245} & \textbf{0.1558} & \underline{0.6842} & \underline{0.6361}  \\
    \bottomrule
    \end{tabular}
}
\vspace{-0.8em}
\label{table:model_ablation}
\end{table}

\subsection{Ablation Study (RQ4)}
To assess the importance of the design in CoDiffMob, we perform an ablation study on both datasets, specifically by removing noise prior and the noise fusion process from the model framework. Without noise priors, the model generates trajectories based on initial noise sampled from an i.i.d Gaussian distribution. Conversely, without noise fusion, the model's generation process is informed only by the collaborative transition sampling process, leading to a deficiency in modeling individual behaviors.

Table \ref{table:model_ablation} presents the experimental results, showing a significant performance drop in flow-related metrics when the model operates without noise priors. The model without noise fusion performs best on flow-related metrics but exhibits a substantial discrepancy from the real data in terms of modeling individual movement behaviors, such as the distribution of individual activity radius and the number of daily visited locations. By combining informative noise based on the collaborative sampling process with white noise, our model can capture both individual movement characteristics and crowd movement behaviors.

\section{Related Works}
\subsection{Noise Prior in Video and Image Generation}
Diffusion models~\cite{luo2023videofusion, yang2023diffusion} have made significant progress in video generation tasks, but ensuring high-quality frame-to-frame consistency remains a core challenge~\cite{croitoru2023diffusion_survey}. Recent studies~\cite{ge2023preserve, zhang2024trip, qiu2023freenoise,sheng2025unveiling} have addressed this by enhancing content coherence through strategic noise manipulation during the diffusion process. For example, PYoCo~\cite{ge2023preserve} proposes a novel noise sampling method preserving correlations between frames, and TRIP~\cite{zhang2024trip} alters the noise targets predicted by the model during the training process. For image generation, Narek \textit{et al.}~\cite{tumanyan2023plug} preserves the structural information of images through noise priors, enabling style control based on the original image structure.
However, unlike video and image data, which primarily focuses on temporal consistency and content information, human mobility data presents more complex patterns and higher levels of stochasticity~\cite{luca2021survey}. 
Designing effective noise priors for such data remains an open research challenge, but it holds immense potential. This paper pioneers the exploration of noise priors in the context of urban mobility generation.

\subsection{Mobility Generation}
Urban mobility trajectory generation methods can broadly be divided into two categories: non-generative and generative approaches. Non-generative methods, such as Markov-based models~\cite{jiang2016timegeo, song2010modelling,bindschaedler2016synthesizing} and simulation techniques~\cite{feng2020learning,isaacman2012human}, rely heavily on historical data, predefined rules, or statistical models to synthesize mobility patterns. However, their reliance on fixed assumptions limits their ability to capture the complexity and variability inherent in human mobility dynamics.
Generative approaches, on the other hand, leverage generative AI techniques~\cite{goodfellow2020generative,ho2020ddpm,kingma2013auto} to model and create synthetic human trajectories with greater complexity and diversity. Early generative efforts, such as Generative Adversarial Networks (GANs)~\cite{goodfellow2020generative}, Generative Adversarial Imitation Learning (GAIL)~\cite{ho2016gail}, and Variational Autoencoders (VAEs)~\cite{kingma2013auto}, pushed the field forward but often struggled with training stability and model collapse. Recent advancements have introduced diffusion models~\cite{ho2020ddpm, song2020ddim, song2020score} for trajectory generation~\cite{zhu2023difftraj, chu2024simulating}, representing a promising new direction in generative modeling. Diffusion-based models offer greater flexibility and diversity, enabling more accurate and nuanced human mobility simulations. We followed the common approach in diffusion research of using UNet~\cite{ronneberger2015unet} to model data correlations, employing CNN and attention mechanisms to capture the spatiotemporal associations in trajectory data. Additionally, we innovatively informed the generation of individual trajectories by incorporating collective movement patterns in the form of noise priors.

\section{Discussion and Conclusion}
Supported by vast amounts of user data from the web, data-driven AI methods have significantly enhanced web applications and services, such as personalized recommendations, leading to substantial improvements in user experience.
However, it is crucial to ensure that user privacy is protected throughout the process to maintain an efficient, secure, and ethical web environment.

We address this challenge from the perspective of data generation. CoDiffMob focuses on generating high-quality synthetic individual mobility trajectory data.
An essential question here is how reliable our proposed deep model is in generating synthetic data while safeguarding privacy. Because empirical studies have shown the potential for identifying individuals through the uniqueness of their mobility trajectories~\cite{tu2018protecting, xu2017trajectory}. Comprehensive privacy evaluations in our work confirm that our model does not leak user privacy, contributing to more secure web services.

From a practical standpoint, aligning with the UN SDGs for sustainable city development, numerous efforts have been dedicated to promoting equitable and sustainable urban planning~\cite{zheng2023spatial, zheng2023road, su2024metrognn}.
These efforts address land use planning~\cite{zheng2023spatial}, road infrastructure~\cite{zheng2023road}, and public transportation management~\cite{su2024metrognn}, focusing on urban sustainability, transportation resilience, and equitable access to green spaces—factors directly impacting the well-being of city residents.  However, existing research often explores optimization algorithms without a robust simulation of users' movements, which may lead to a disconnect between theoretical objectives and residents' actual movement needs.
Our approach, by generating high-quality urban mobility data, provides a powerful tool for advanced urban decision-making~\cite{zheng2024survey}. It supports more informed planning and optimization, enabling the design of more resilient, sustainable, and equitable urban environments.

\begin{acks}
This work was supported in part by the National Key Research and Development Program of China under grant 2023YFB2904804, the National Nature Science Foundation of China under 62476152 and U24B20180, and Beijing National Research Center for Information Science and Technology (BNRist).
\end{acks}

\bibliographystyle{ACM-Reference-Format}
\balance
\bibliography{sample-base}

\appendix

\section{Methodology Details}

\subsection{Diffusion Denoiser}
Following ~\cite{zhu2023difftraj}, we apply a UNet-based diffusion denoiser to generate individual trajectories from Gaussian noise.
During the network computation, the noised trajectory is represented as a $ T \times C $ matrix, where $ T $ denotes the length of the trajectory and $ C $ represents the number of dimensions for the location information at each time step (e.g., $ C = 2 $ for latitude and longitude). In the upsampling and downsampling stages of the UNet architecture, one-dimensional convolutional neural networks (CNNs) are employed to model the spatiotemporal correlations between trajectory points. Additionally, an attention mechanism is applied to the features after downsampling, allowing the model to capture complex relationships within the latent space. The hyper-parameters used to build the model are shown in Table~\ref{table:para_denoiser}.

\begin{table}[htbp]
\centering
\caption{Hyper-parameters of the diffusion denoiser.}
\resizebox{0.5\linewidth}{!}{
    \begin{tabular}{lc}
    \toprule
    Parameter & Value  \\ \midrule
    Input Channel Size & 2 \\
    Output Channel Size & 2 \\
    Trajectory Length & 48 \\ \midrule
    Num Freq Bands & 64 \\
    Hidden Dim & 128 \\
    Num Initial Channel & 128 \\
    Channel Mult & [1, 2, 2, 2] \\
    Channel Mult Emb & 4 \\
    Channels Per Head & 64 \\
    Num Blocks & 2 \\
    \midrule
    Batch Size & 1024 \\
    Guidance Scale & 3 \\
    \bottomrule
    \end{tabular}
}
\label{table:para_denoiser}
\end{table}

\begin{figure*}[t]
\begin{center}
\includegraphics[width=0.8\linewidth]{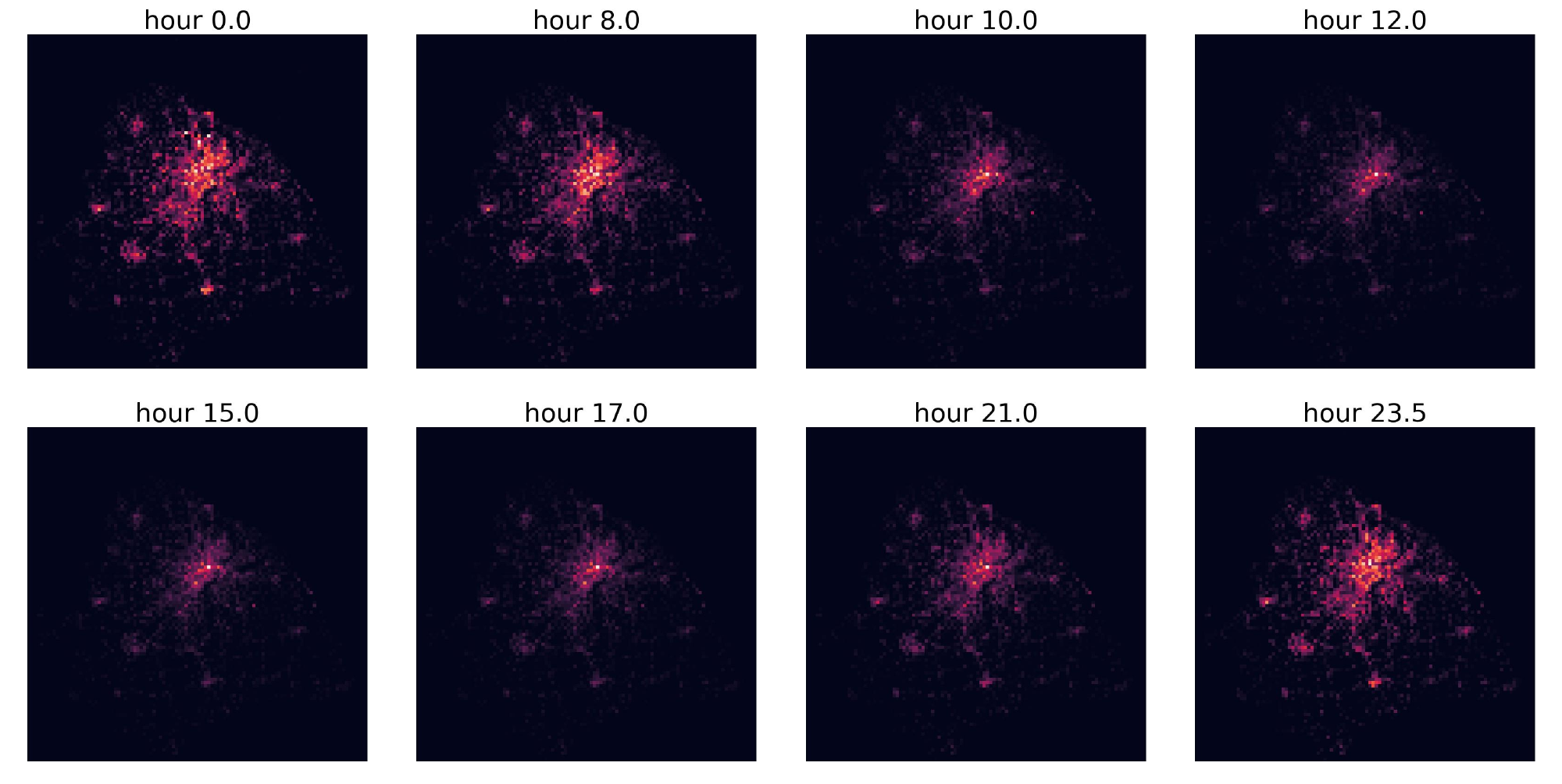}
\end{center}
\caption{Visualization of the trajectory distribution of ISP dataset at different timestamps.}
\label{fig:isp_mobility}
\end{figure*}

\subsection{Training Details}
\label{appendix:train}
Diffusion model estimates the distribution of the real-world data $\boldsymbol{x} \sim p(\boldsymbol{x})$ by sampling from the approximate distribution $p_{\boldsymbol{\theta}}(\boldsymbol{x})$ with learnable parameter $\boldsymbol{\theta}$. 
Normally we denote $p_{\boldsymbol{\theta}} (\boldsymbol{x})= \frac{ - f_{\boldsymbol{\theta}} (\boldsymbol{x})}{Z_{\boldsymbol{\theta}}}$ , and optimize the parameter $\boldsymbol{\theta}$ by max-likelihood $\max_{\boldsymbol{\theta}} \sum_{i=1}^{N} \log{p_{\boldsymbol{\theta}}(\boldsymbol{x}_i)}$. 
However, we need to know the normalization constant $Z_{\boldsymbol{\theta}}$ to make the max-likelihood training feasible, and approximating it would be a very computationally expensive process. Therefore, we choose to model the score function $\nabla_{\boldsymbol{x}} \log{p_{\boldsymbol{\theta}}(\boldsymbol{x};\sigma)}$ rather than the probability density, with the score function, we are able to get data sample $\boldsymbol{x}_0 \sim p_{\boldsymbol{\theta}}(\boldsymbol{x})$ by the following equation:

\begin{equation}
\label{eq:diff-sampling}
\begin{array}{l}
\boldsymbol{x}_{0}=\boldsymbol{x}(T)+\int_{T}^{0}-\dot{\sigma}(t) \sigma(t) \nabla_{\boldsymbol{x}} \log p_{\boldsymbol{\theta}}(\boldsymbol{x}(t) ; \sigma(t)) d t,
\vspace{0.5em} \\
\text {where }  \boldsymbol{x}(T) \sim \mathcal{N}\left(\mathbf{0}, \sigma_{\max }^{2} \boldsymbol{I}\right).
\end{array}
\end{equation}

\noindent On this basis, we add a condition $\boldsymbol{c}$ composed of the urban environment and use our model to approximate the score function $\nabla_{\boldsymbol{x}} \log p_{\boldsymbol{\theta}}(\boldsymbol{x}; \boldsymbol{c}, \sigma) \approx \left(D_{\boldsymbol{\theta}}(\boldsymbol{x}; \boldsymbol{c}, \sigma)-\boldsymbol{x}\right) / \sigma^{2}$, which leads to the training target:

\begin{equation}
\label{eq:diff-train}
\mathcal{L}(\theta) = \mathbb{E}_{\boldsymbol{x}, \boldsymbol{c} \sim \chi_{c}} \mathbb{E}_{\sigma \sim q(\sigma)} \mathbb{E}_{\boldsymbol{\epsilon} \sim \mathcal{N}\left(\mathbf{0}, \sigma^{2} \boldsymbol{I}\right)}\left\|D_{\boldsymbol{\theta}}(\boldsymbol{x}+\boldsymbol{\epsilon} ; \boldsymbol{c}, \sigma)-\boldsymbol{x}\right\|_{2}^{2},
\end{equation}

\noindent here $\chi_{c}$ is the training dataset combined with embeddings computed by the scene encoder, and $q(\sigma)$ represents the schedule of the noise level added to the original data sample.
For better performance, we introduce the precondition as described in~\cite{karras2022elucidatingdesignspacediffusionbased} to ensure that the input and output of the model both follow a standard normal distribution with unit variance:

\begin{equation}
\label{eq:diff-precondition}
D_{\boldsymbol{\theta}}(\boldsymbol{x} ; \boldsymbol{c}, \sigma)=c_{\text {skip }}(\sigma) \boldsymbol{x}+c_{\text {out }}(\sigma) F_{\boldsymbol{\theta}}\left(c_{\text {in }}(\sigma) \boldsymbol{x} ; \boldsymbol{c}, c_{\text {noise }}(\sigma)\right),
\end{equation}

\noindent here $F_{\boldsymbol{\theta}}(\cdot)$ represents the original output of the diffusion denoiser. In the experiment, we use the longitude and latitude of trajectory points as the generation target. Details of the training hyper-parameters can be found in Table~\ref{table:para_train}.




\begin{table}[htbp]
\centering
\caption{Hyper-parameters of the training process.}
\resizebox{0.72\linewidth}{!}{
    \begin{tabular}{lc}
    \toprule
    Parameter & Value  \\ \midrule
    Num Epochs & 200 \\
    Weight Decay & 0.03 \\
    Learning Rate & 0.0005 \\
    Learning Rate Schedule & OneCycleLR \\
    $\sigma_{data}$ & 0.1 \\ \vspace{0.2em}
    $c_{in}(\sigma)$ & $ 1 / \sqrt{\sigma^2 + \sigma_{data}^2}$ \\ \vspace{0.2em}
    $c_{skip}(\sigma)$ & $ \sigma_{data}^2 /( \sigma^2 + \sigma_{data}^2 )$ \\ \vspace{0.2em}
    $c_{out}(\sigma)$ & $ \sigma \cdot \sigma_{data} / \sqrt{\sigma^2 + \sigma_{data}^2}$ \\ \vspace{0.2em}
    $c_{noise}(\sigma)$ & $ \frac{1}{4} \ln{\sigma}$ \\ \vspace{0.2em}
    Noise Distribution & $ \ln (\sigma) \sim \mathcal{N}\left(P_{\text {mean }}, P_{\text {std }}^{2}\right)$ \\
    $P_{\text {mean }}$ & -1.2 \\
    $P_{\text {std }}$ & 1.2 \\
    \bottomrule
    \end{tabular}
}
\label{table:para_train}
\end{table}

\begin{table}[htbp]
\centering
\caption{Statistics of the two mobility datasets.}
\resizebox{0.72\linewidth}{!}{
    \begin{tabular}{lccccc}
    \toprule
    Dataset & City & Duration & \#Users & \#Loc & \#Traj \\ 
    \midrule
    ISP & Shanghai & 7 days & 90037 & 9158 & 261042 \\
    MME & Nanchang & 7 days & 6218 & 4096 & 43967 \\
    \bottomrule
    \end{tabular}
}
\label{table:datasets}
\end{table}

\section{Experiment Details}
\subsection{Datasets}
\label{appendix:dataset}
We conduct experiments on two real-world mobility datasets collected based on web applications and services. The statistics of the two datasets are summarized in Table~\ref{table:datasets}.

\begin{itemize}[leftmargin=*]
    \item \textbf{ISP}: The ISP dataset is collected through a partnership with a major Internet Service Provider (ISP) in China. The mobility data contains more than two million users, with a primary spatial scope of Shanghai from April 19 to April 26, 2016. Each mobility record in the ISP dataset contains an anonymous user ID, timestamp, and cellular base station.
    \item \textbf{MME}:  The MME dataset is provided by an operator in China. The dataset contains the trajectories of more than 6000 users over a week in Nanchang, with each mobility record containing an anonymous user ID, time stamp, and region ID.
\end{itemize}

Figure~\ref{fig:isp_mobility} illustrates the distribution of user trajectory points over time throughout the day in the ISP dataset. 

\begin{figure*}[htbp]
\begin{center}
\includegraphics[width=0.75\linewidth]{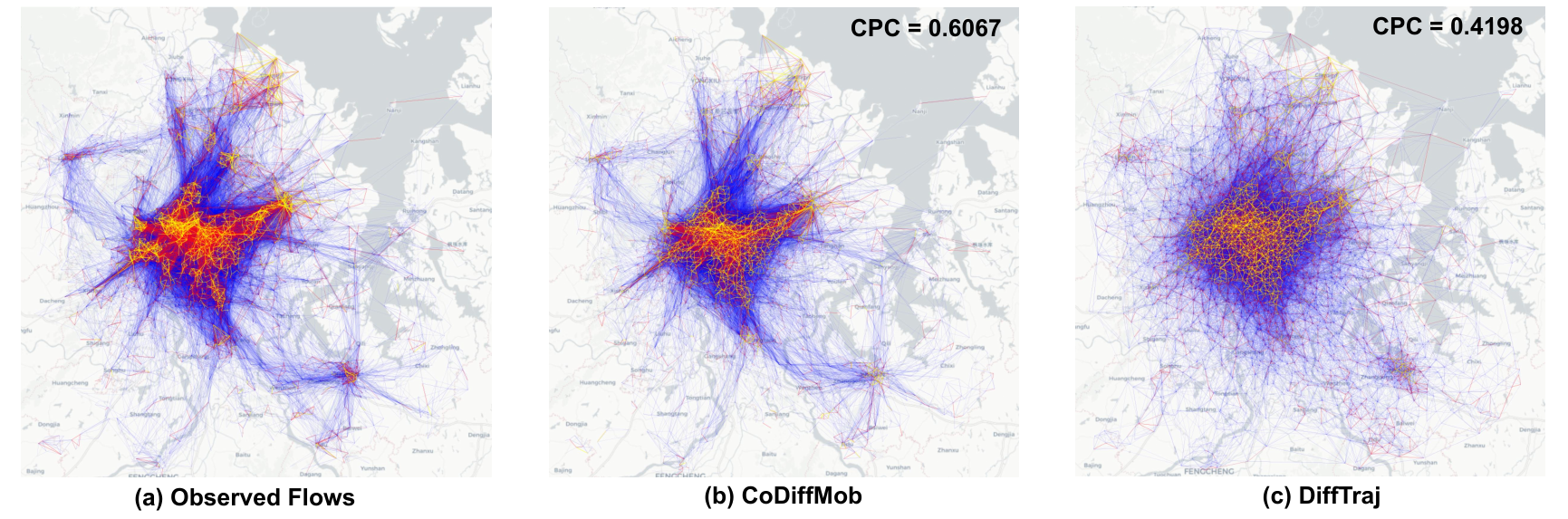}
\end{center}
\caption{Visualization of the real and generated flow on MME dataset.}
\vspace{1em}
\label{fig:flow_vis_MME}
\end{figure*}

\subsection{Metrics}
\label{appendix:metrics}
We evaluate the generated results of our method from two aspects, individual behavior and collective movement patterns. For individual behavior, we evaluate the generated trajectories from the following four terms:

\begin{itemize}[leftmargin=*]
    \item \textbf{Radius}: Radius tracks the radius of gyration centered in the trajectory's center of mass, we calculate the radius for each trajectory with a length of one day.
    \item \textbf{Distance}: Distance measures the travel distance for each move in the trajectory, only calculated when a user moves from one location to another.
    \item \textbf{Duration}: Duration calculates the time spent at each location, we excluded trajectories that remained at the same location throughout the entire day in our calculations.
    \item \textbf{DailyLoc}: DailyLoc records the number of locations visited by each person on a daily basis. Similarly, trajectories that remained at the same location throughout the day are not considered.
\end{itemize}

We do Kolmogorov–Smirnov Test~\cite{massey1951kolmogorov} between the metrics calculated from the real trajectories and the generated ones. For a metric $X$, the empirical distribution function $F_n$ with $n$ samples is defined as:

\begin{equation}
\label{eq:dis_func}
\begin{array}{l}
F_n(x) = \frac{\text{number of }(\text{elements in sample }\leq x)}{n} 
\vspace{0.5em}\\
\quad \quad \, \, \, \, = \frac{1}{n}\sum_{i=1}^n\mathbb{I}_{(-\inf,x]}(X_i),
\end{array}
\end{equation}

\noindent and with the empirical distribution function of the real samples $F_{1,n}$ and generated samples $F_{2,m}$, the Kolmogorov–Smirnov statistic is defined as:

\begin{equation}
\label{eq:dis_func}
\begin{array}{l}
D_{n,m} = \sup_x \|F_{1,n}(x) -  F_{2,m}(x)\|.
\end{array}
\end{equation}

For generated flows, following previous work \cite{simini2021deep}, we calculate \textbf{Common Part of Commuters} (CPC) and \textbf{MAPE} between the real transition matrix and those inferred from the generated data. Given real-world dataset $\mathcal{X}$ and generated data $\mathcal{Y}$, we calculate collective flows  $\mathcal{F}^{\mathcal{X}}$ and $\mathcal{F}^{\mathcal{Y}}$ from them, CPC is calculated by the following formula:

\begin{equation}
\label{eq:cpc}
\begin{array}{l}
NCC(\mathcal{F}^{\mathcal{X}},\mathcal{F}^{\mathcal{Y}}) = \sum_{i=1}^n\sum_{j=1}^n\min(\mathcal{F}^{\mathcal{X}},\mathcal{F}^{\mathcal{Y}}), 
\vspace{0.5em} \\
NC(\mathcal{F}^{\mathcal{X}}) = \sum_{i=1}^n\sum_{j=1}^n\mathcal{F}^{\mathcal{X}}, 
\vspace{0.5em} \\
CPC(\mathcal{F}^{\mathcal{X}},\mathcal{F}^{\mathcal{Y}}) = \frac{2NCC(\mathcal{F}^{\mathcal{X}},\mathcal{F}^{\mathcal{Y}})}{NC(\mathcal{F}^{\mathcal{X}}) + NC(\mathcal{F}^{\mathcal{Y}})},
\end{array}
\end{equation}

\noindent MAPE is calculated by the average percentage error of the real and generated transition probability of each location $\widetilde{\mathcal{F}}_{l_{i}}^{X}$ and $\widetilde{\mathcal{F}}_{l_{i}}^{Y}$:

\begin{equation}
\label{eq:cpc}
\begin{array}{l}
MAPE = \frac{1}{n}\sum_{i=1}^n\|\frac{\widetilde{\mathcal{F}}_{l_{i}}^{\mathcal{X}} - \widetilde{\mathcal{F}}_{l_{i}}^{\mathcal{Y}}}{\widetilde{\mathcal{F}}_{l_{i}}^{\mathcal{X}}}\|,
\end{array}
\end{equation}

\noindent here We filtered out data points with real transition probabilities lower than $0.01$ to ensure the validity of the results, and we used the $l_1$ norm to measure the difference between the two transition probability densities.

Also, we introduce a \textbf{Diversity} metric, which calculates the proportion of trajectories generated by the model that are identical to those appearing in the real dataset:

\begin{equation}
\label{eq:diversity}
\text{Diversity}(\mathcal{Y}) = \frac{\sum_{\boldsymbol{y} \in \mathcal{Y}} \mathbb{I}(\boldsymbol{y} \in \mathcal{X})}{|\mathcal{Y}|},
\end{equation}

\noindent here $\mathcal{X}$ represents the real-world dataset and $\mathcal{Y}$ is the generated one.

\subsection{Baselines}
\label{appendix:baselines}
We compare the performance of CoDiffMob with six state-of-the-art baselines for mobility generation:
\begin{itemize}[leftmargin=*]
    \item \textbf{TimeGEO} \cite{jiang2016timegeo}: A model-based trajectory synthesis method that models temporal decisions using hyperparameters while employing the EPR model~\cite{song2010modelling} to capture spatial decision-making.
    \item \textbf{PateGail} \cite{wang2023pategail}: A robust generative adversarial imitation learning model to mimic the human decision-making process.
    \item \textbf{MoveSim} \cite{feng2020learning}: A generative adversarial framework that uses insights from human mobility regularities for mobility generation.
    \item \textbf{VOLUNTEER} \cite{long2023practical}: It utilizes a two-layered VAE to capture the joint distribution of user attributes and mobility behaviors.
    \item \textbf{TrajGDM} \cite{chu2024simulating}: A trajectory generation framework built on the diffusion model, aiming to capture common mobility patterns within a trajectory dataset.
    \item \textbf{DiffTraj} \cite{zhu2023difftraj}: A UNet-based diffusion model to capture the spatiotemporal features of individual movements.
\end{itemize}

\section{Additional Results}

\begin{table}[htbp]
\centering
\caption{Results of the ablation study on MME dataset.}
\resizebox{1.0\linewidth}{!}{
    \begin{tabular}{lcccccc}
    \toprule
    \multirow{2}{*}{} & \multicolumn{4}{c}{Trajectory} & \multicolumn{2}{c}{Flow} \\ 
    \cmidrule(l){2-5} 
    \cmidrule(l){6-7}   
    & Radius & Distance & Duration & DailyLoc & CPC    & MAPE    \\ 
    \midrule
    w/o Noise Prior      & \underline{0.1010} & \underline{0.1250} & 0.0940 & \underline{0.2656} & 0.4061  & 0.7912  \\
    w/o Noise Fusion    & 0.2298 & 0.1484 & \underline{0.0861} & 0.3898 & \textbf{0.6910}  & \textbf{0.5609}  \\
    CoDiffMob         & \textbf{0.0519} & \textbf{0.0970} & \textbf{0.0796} & \textbf{0.1729} & \underline{0.6067} & \underline{0.6731}  \\
    \bottomrule
    \end{tabular}
}
\label{table:model_ablation_mme}
\end{table}

Figure~\ref{fig:flow_vis_MME} presents the visualization of real and generated collective flow data based on the MME dataset. Compared to the best baseline, DiffTraj, our method is shown to better capture the collective movement patterns in the real data.
Table~\ref{table:model_ablation_mme} presents the results of the ablation study on the MME dataset. The findings align with the conclusions drawn from the ISP dataset, as discussed in the main text. Specifically, the noise prior improves model performance at both the individual behavior and collective movement levels. Moreover, the noise fusion design effectively integrates rule-based collective movement patterns with model-parameter-learned individual behavior characteristics, enabling the generated results to balance performance across both individual and collective levels.

\end{document}